\documentclass[runningheads]{llncs}
\usepackage{graphicx}
\usepackage{color}
\begin{document}
\pagestyle{empty}
\title{Efficient Single Image Super Resolution using Enhanced Learned Group Convolutions}
\author{Vandit Jain \and
Prakhar Bansal \and
Abhinav Kumar Singh \and
Rajeev Srivastava}
\institute{Indian Institute of Technology (Banaras Hindu University), Varanasi \linebreak
\{vandit.jain.cse15, prakhar.bansal.cse15, abhinavkr.singh.cse15, rajeev.cse\}@ iitbhu.ac.in}

\maketitle              
\begin{abstract}
Convolutional Neural Networks (CNNs) have demonstrated great results for the single-image super-resolution (SISR) problem. Currently, most CNN algorithms promote deep and computationally expensive models to solve SISR. However, we propose a novel SISR method that uses relatively less number of computations. On training, we get group convolutions that have unused connections removed. We have refined this system specifically for the task at hand by removing unnecessary modules from original CondenseNet. Further, a reconstruction network consisting of deconvolutional layers has been used in order to upscale to high resolution. All these steps significantly reduce the number of computations required at testing time. Along with this, bicubic upsampled input is added to the network output for easier learning. Our model is named SRCondenseNet. We evaluate the method using various benchmark datasets and show that it performs favourably against the state-of-the-art methods in terms of both accuracy and number of computations required.

\end{abstract}
\section{Introduction}
Super Resolution (SR) problem is defined as recovering a high resolution image from a low resolution image. This is a highly ill-posed problem with multiple solutions possible for a single input image. This problem finds many applications such as medical imaging, security and surveillance among others.

In recent years, deep learning methods have performed better as compared to interpolation-based\cite{Duchon}, reconstruction-based \cite{glasner2009,Yang2010}  or other example-based methods \cite{Schulter2015,A+,Perez,Salvador2015} that have been used in the past. This is proved by the fact that the first effort in the direction of deep learning for solving the problem of single image super resolution \cite{SRCNN} performed better than several previous models not using deep learning algorithms.

This lead to development of several other methods that used deep learning\cite{VDSR,DRCN,SRResNet,EnhanceNet,SRDenseNet,EDSR,SESR,LapSRN,DRRN}. However, all these methods in order to get a slight performance improvement (in terms of PSNR) promoted use of deep, computationally heavy CNN models. It would be objectively correct to say that such heavy resources are not available at all situations for such lengthy periods of time. In order to solve this problem, it is required to build a model that uses less number of multiplication-addition operations (FLOPs) to come up with a high resolution image.

In this work, we present a novel super resolution model termed SRCondeseNet that uses the concept of removing unused connections in the network to form group convolutions. Normal group convolutions also help in reducing the number of connections but the latter method comes with a huge loss in accuracy. Once the features are extracted using this reduced model, reconstruction is done using deconvolutional layers with 1x1 kernels to produce a high resolution image. Also applied is the concept of residual learning i.e. the bicubic upsampled input is added to network output so that the model only has to learn the difference\cite{VDSR}. Our contributions through this work are:
\begin{enumerate}
    \item Our model incorporates the use of group convolutions and pruning in the field of super resolution thereby producing a lightweight CNN model for this problem.
    \item Setting state-of-the-art in terms of performance metrics such as PSNR and SSIM along with using less number of FLOPs as compared to current light weight SISR methods.  
\end{enumerate}

\section{Related Work}
Here we focus on various deep learning methods that have been used to solve the SISR problem. Also we go through various methods that have been proposed to come up with efficient, lightweight CNNs.

\subsection{Single Image Super Resolution}
Various deep learning methods have been applied in the past, to solve the SISR problem, many of which have been summarized in \cite{Hayat}. First, Dong et al. proposed in \cite{SRCNN} the replacement of all steps to produce a high resolution image - feature extraction then mapping  then reconstruction - by a single neural network. The deep learning model performed better than other example-based methods. However, it was proposed in \cite{SRCNN} that deeper networks may not be effective for SISR. This was proved wrong by Kim et al. in \cite{VDSR}. They used a very deep CNN model that performed better than \cite{SRCNN}. Kim et al. in \cite{VDSR} used residual learning proposed by He et al. in \cite{He2016DeepRL} to combat the problem of vanishing gradients that arises in deep models. Since then, the concept of residual learning has been used by many CNN models\cite{SRResNet,SRDenseNet,SESR,LapSRN,DRRN,MemNet}. Hence, we also include the feature of global residual learning in order to avoid the vanishing gradient problem that is bound to happen in a deep model like ours. Moreover, some models\cite{DRCN,SESR,DRRN,MemNet} advocate the use of recursive layers in the CNN. This helps in reducing the number of effective parameters required. However, these models require heavy computation power, significantly higher than our model. To achieve real time performance, \cite{ESPCN} proposed use of sub-pixel convolutions instead of bicubic upsampled image taken as the input. Similarly, \cite{FSRCNN,SRDenseNet,LapSRN} start with low-resolution image as input to the network. This way the model works on low resolution image thereby helping in reducing the number of computations. We also use this idea for the aforementioned purpose. Some methods\cite{SRResNet,EnhanceNet} have advocated the use of GAN (Generative Adverserial Networks) to produce visually-pleasing images along with promising results on quantitative metrics like PSNR and SSIM.

\subsection{Efficient convolutional neural networks}
Many attempts have been made to build CNNs that use less computation power without compromising accuracy. One such method is weight pruning. Weight pruning is removing unwanted connections in a neural network. CondenseNets\cite{CondenseNet} which are explained below use weight pruning. 

\iffalse 
Some notable examples include \cite{MobileNet,ShuffleNet,SqueezeNet}. First two use depth-wise separable convolutions whereas\cite{SqueezeNet} works by reducing number of connections to computationally expensive 3x3 layers. On the other hand, weight pruning is done in some methods\cite{CondenseNet,LiuSlimming,LiPruning}. Weight pruning is removing unwanted connections in a neural network.\cite{LiuSlimming} uses the method of channel sparsity regularisation to prune channels.\cite{LiPruning} instead of removing certain weights removes selected filters with all its feature maps as a whole. This reduces required computational power significantly. All these models share a key idea: getting a quantitative method to choose which are unwanted connections and remove them.

\fi

\subsubsection{CondenseNet}
\begin{figure}
\centering
\includegraphics[scale = 0.35]{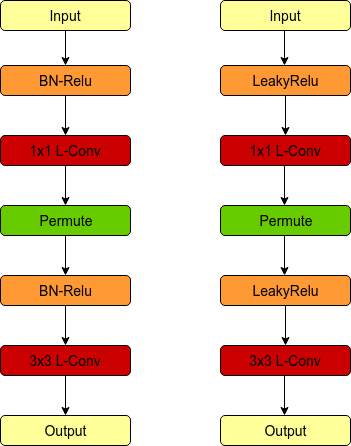}
\caption{Comparison between structure of $denselayer$ of CondenseNet\cite{CondenseNet} (left) and SRCondenseNet (right). BN layer is removed and Relu is replaced by LeakyRelu.} \label{fig1}
\end{figure}

Our model employs blocks that are modified version of  CondenseNet blocks\cite{CondenseNet}. Original CondenseNet blocks use learned group convolutions. In this method, the model goes through two kinds of stages: condensing stages and optimization stage. In the former, using sparsity inducing regularization, unimportant filters are removed. The convolutional layers used here have 1x1 kernels. Thus, number of connections depend on number of input channels and number of output channels only. The condensation is done by calculating $L_1$-norm over every incoming feature and every filter group. Then we remove those columns that have $L_1$-norm value lesser than other columns. The number of feature map connections that are left after every condensing stage depends on the condensing factor $C$. Once we get the lighter model, it goes through optimization stage where it is trained. Every block contains several $denselayers$  and structure of each original $denselayer$ is described in Figure:~\ref{fig1} (left).

\section{Proposed Method}
In this section, we describe our proposed method, SRCondenseNet, in detail. First we take as input the low resolution(LR) image and pass it into an input convolutional layer. The output of this layer is fed into modified CondenseNet that contains $denselayers$ that are stacked into four blocks. The output of the last block is sent to what is called as the reconstruction network. It comprises of a bottleneck layer, a set of deconvolution layer, whose number depends on the scaling factor. Next comes the reconstruction layer with one output channel to get the final image.

\subsection{Modified CondenseNet blocks}
In section 2.2, we explained original CondenseNet blocks. However, CondenseNet has been designed for classification task. Hence, several modifications have been done to suit it for the SISR problem. SRCondenseNet contains $denselayer$ structure, which has been depicted in Figure:~\ref{fig1} (right).

\begin{figure}
\centering
\includegraphics[scale = 0.48]{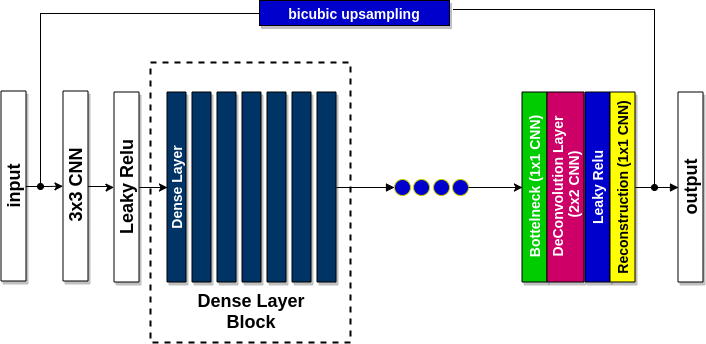}
\caption{Our model (SRCondenseNet) structure. We have four such blocks(black dashed line) each containing seven $denselayer$ (blue) structures. Only one block is shown for clarity. This is followed by the reconstruction network containing bottleneck, deconvolution and reconstruction CNN at the end.} \label{fig2}
\end{figure}

Every block contains many $denselayer$ named structures. We have removed Batch-Normalization layers as suggested by Nah et al.\cite{Nah2016} and Lim et al.\cite{EDSR}. This removes unnecessary computations. Also Relu activation has been replaced by LeakyRelu to combat the ``dying ReLU" problem. We stack up four blocks each containing 7 $denselayers$ (blue) as shown in Figure:~\ref{fig2}. Only one out of four blocks (black dashed line) is shown in the figure to avoid clutter. Number of input channels in every $denselayer$ depends on growth rate and increase in number linearly according to it. Every block has its own growth rate. After testing several values for trade-off between model size and accuracy, we set growth rate of all the blocks to 20. Thus, every subsequent $denselayer$ has 20 more input channels than the previous one.

Moreover, original CondenseNet contains transition layers between blocks comprising of average pooling layers. For SISR problem, there was no need of pooling layers. We have skipped these transition layers in our model. Thus, width and height of input into first block is equal to width and height of output from last block.

\subsection{Reconstruction Network}
The reconstruction network comes after the modified CondenseNet blocks as shown in Figure ~\ref{fig2}. It starts with the bottleneck layer (green) which is a 1x1 layer to reduce the output feature maps to a very less number thereby reducing the number of computations in further layers. 1x1 kernel also helps in the purpose. Number of output feature maps are set to 128.
Next, this is followed by a set of deconvolutional layers (pink). Their number depends on the scaling factor$(r)$. With r equal to 2, we have a single deconvolutional layer with stride 2. Deconvolutional layers help in reducing the number of parameters and computational complexity by a factor of $r^{2}$ throughout the model. This is because, by using bicubic interpolated image as an input, instead of upscaling it at the last using deconvolutional layers, increases the size of input to all feature extraction layers by a factor of $r^{2}$. This method of upscaling also improves performance in reconstruction. Again, number of feature maps are set to 128 for all deconvolutional layers. Finally, we end up the model with a convolutional layer (yellow) with one channel as output to get the final YCbCr image.

\subsection{Global Residual Learning}
Deep CNN models with high number of layers tend to suffer from vanishing gradient problem. Hence, as proposed by He et al. in \cite{He2016DeepRL}, this problem is solved by adding a global residual connection. In our model, we add a residual connection in which we add a bicubic interpolated image to the output received from the model. This makes the learning easier and more and more layers can be stacked.

\section{Experiments}
\subsection{Datasets}
We have used 91 images from Yang et al.\cite{Yang91} and 200 images from the Berkeley Segmentation Dataset(BSD)\cite{BSD200} for training. We cut out several patches of the original images with a stride of 64, size of which depends on the scaling factor. For every case, during training, input size of the image to the network is 32x32. Hence, for scaling factor of 2, we cut out patches of 64x64.  Further, we have performed data augmentation on these patches. Eventually, we get five patches for a single original patch. These are converted to YCbCr image and only Y-channel is processed.

We test our model on standard datasets: SET5\cite{set5}, SET14\cite{set14} and Urban100 \cite{urban100}. 

\subsection{Implementation Details}
We set the initial learning rate to 0.0001 and keep the cosine learning rate method as used by Huang et al. in \cite{CondenseNet}. We run the network for 180 epochs with both condensing factor and number of groups set to 4 to have every condensing stage with 30 epochs. LeakyReLUs have negative slope set to $0.1$. We train our network on a Tesla P40 GPU. All networks were optimized using Adam\cite{adam}. We have used a robust Charbonnier loss function instead of $L1$ or $L2$ function that is generally used\cite{SRCNN,VDSR,DRCN,EDSR} to aid high-quality reconstruction performance\cite{LapSRN}.  

\subsection{Comparison with State-of-the-Art Methods}
\subsubsection{Comparison on the basis of accuracy}
\begin{figure}
\centering
\includegraphics[scale = 0.5]{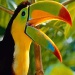}
\includegraphics[scale = 0.5]{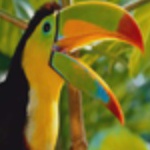}
\includegraphics[scale = 0.5]{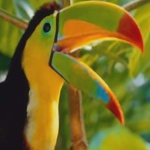}
\includegraphics[scale = 0.5]{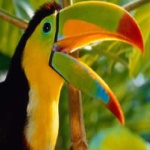}
\linebreak
\includegraphics[scale = 0.5]{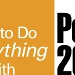}
\includegraphics[scale = 0.5]{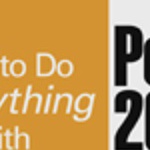}
\includegraphics[scale = 0.5]{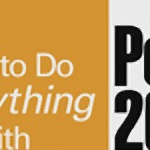}
\includegraphics[scale = 0.5]{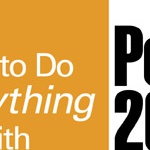}
\caption{Qualitative comparison. The first row shows low resolution input, bicubic interpolation of LR input, output from our model, original HR image(left to right) of image from set5. Similarly second row are images for img\_013 from set14. } \label{fig3}
\end{figure}

Peak signal-to-noise Ratio (PSNR) and structures similarity (SSIM) are the two standard metrics for comparison. Table~\ref{tab:tab1} shows comparisons for SISR results for various models(scale = x2). Clearly, our method performs handsomely when compared to current state-of-the-art models using similar computation power. Various standard testing datasets have been used. Figure~\ref{fig3} shows a qualitative comparison of images from various testing datasets. 

\begin{table}
    \centering
    \caption{Average PSNR/SSIM values for x2 scale factor for various models on different models. \textcolor{red}{red} indicates best value and \textcolor{blue}{blue} indicates second best value.}
    \label{tab:tab1}
    \begin{tabular}{|c|c|c|c|c|c|}
        \hline
    Dataset & SRCNN\cite{SRCNN} & VDSR\cite{VDSR} & LapSRN\cite{LapSRN} & DRRN\cite{DRRN} & SRCondenseNet(ours) \\
         \hline
        Set5 & 36.66/0.9542 & 37.53/0.9587 & 37.52/0.9590 & \textcolor{blue}{37.74}/\textcolor{blue}{0.9591} & \textcolor{red}{37.79}/\textcolor{red}{0.9594}\\
    \hline
    Set14 & 32.45/0.9067 & 33.03/0.9124 & 33.08/0.9130 & \textcolor{red}{33.23}/\textcolor{blue}{0.9136} & \textcolor{red}{33.23}/\textcolor{red}{0.9137}\\
    \hline
    Urban100 & 29.50/0.8946 & 30.76/0.9140 & 30.41/0.9100 & \textcolor{blue}{31.23}/\textcolor{blue}{0.9188} & \textcolor{red}{31.24}/\textcolor{red}{0.9190} \\
    \hline
    \end{tabular}
    
\end{table}

% \begin{figure}
% \begin{center}
    
% \includegraphics[scale = 0.34]{graph.JPG}
% \caption{Graph showing average FLOPs(SET5\cite{set5} scale x2) vs PSNR trade-off.} \label{fig4}
% \end{center}

% \end{figure}
\begin{figure}
\begin{center}
    
\includegraphics[scale = 0.4]{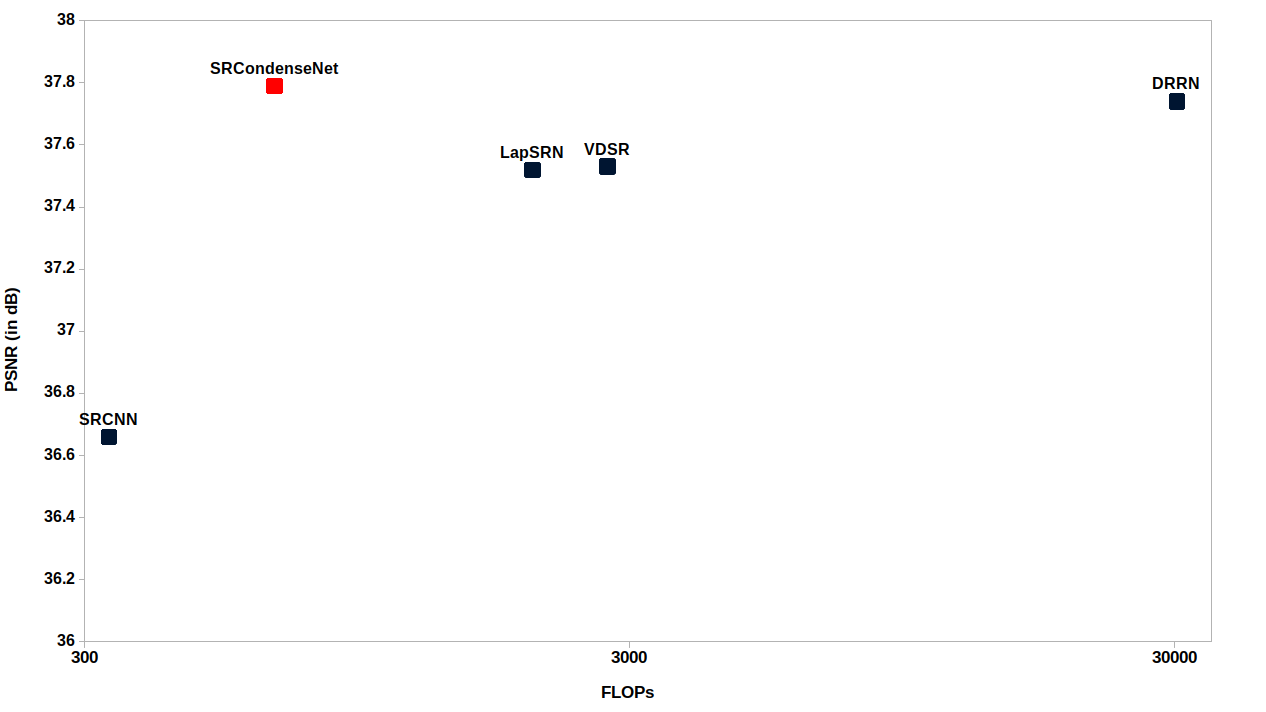}
\caption{Graph showing average PSNR vs FLOPs(SET5\cite{set5} scale x2) trade-off.} \label{fig4}
\end{center}

\end{figure}
\subsubsection{Comparison on the basis of FLOPs}
SRCondenseNet uses the concept of learned group convolutions. Thus, it requires relatively less computation power to produce better results. Here, for comparison, we have used the definition of FLOPs(number of multiplications and additions) to compare computational complexity. Similar method was used in the original CondenseNet \cite{CondenseNet} paper. We have used the same method to calculate FLOPs for all models. Scale is taken as 2 here as well. SRCNN \cite{SRCNN}, VDSR \cite{VDSR} \& DRRN \cite{DRRN} take bicubic interpolated input, hence we take input image size as 64x64 for these models. Whereas, we take 32x32 input image size for LapSRN \cite{LapSRN} \& our model as these models use original low resolution image. Table~\ref{tab:tab2} shows that our model is lighter than most models. There is a trade-off between computational complexity and PSNR as can be seen in figure~\ref{fig4}. SRCNN  \cite{SRCNN} contains only three parameterised convolutional layers and thus is unable to learn good enough mapping between a low resolution image and its coressponding high resolution image. Number of layers in SRCNN \cite{SRCNN} is very less as compared to all other models mentioned in table~\ref{tab:tab1} and table~\ref{tab:tab2} which makes it computationally less expensive (without explicitly applying any technique to reduce number of parameters) than other models (including ours). However, it should also be noted that it produces significantly poorer results than all other models. On the other hand, rest of all the models are computationally heavier than our model.      

\begin{table}
    \centering
    \caption{FLOPs count (x1e6) for various models with suitable input to produce a size 64x64 output image and scale factor of 2. \textcolor{red}{red} indicates best value and \textcolor{blue}{blue} indicates second best value.}
    \label{tab:tab2}
    \begin{tabular}{|c|c|c|c|c|c|c|}
    \hline
 Model & SRCNN\cite{SRCNN} & VDSR\cite{VDSR} & LapSRN\cite{LapSRN} & DRRN\cite{DRRN} & SRCondenseNet(ours) \\
    \hline
    FLOPs(x1e6) & \textcolor{red}{332.32} & 2727.61 & 1988.38 & 30235.17 & \textcolor{blue}{668.88} \\
    \hline
    \end{tabular}
\end{table}

\section{Conclusion and Future Work}
In this paper, we propose a single image super resolution method that uses pruned CNNs to solve the problem using less number of computations. The proposed method outperforms state-of-the-art by a considerable margin in terms of PSNR and SSIM while maintaining less number of FLOPs than comparable methods. Learned group convolutions after our modifications are found to be performing well for the SISR task. This work promotes the use of efficient CNNs that have been used widely in high-level computer vision tasks into low-level vision tasks such as SISR.

In this work, Charbonnier loss has been used throughout the process. We intend on integrating perceptual loss in the proposed method in order to produce visually pleasing images as claimed by Ledig et al. \cite{SRResNet} and Sajjadi et al. \cite{EnhanceNet} in future.

\subsubsection{Acknowledgements}
The authors are grateful to HP Inc. for their support to the Innovations Incubator Program. They are thankful to other stakeholders of this program including Leadership, and Faculty Mentors at IIT-BHU, Drstikona and Nalanda Foundation. Authors are also grateful to Dr. Prasenjit Banerjee, Nalanda Foundation for his mentoring and support


\end{document}